# 3D Path Planning and Obstacle Avoidance Algorithms for Obstacle-Overcoming Robots

Huang Yuanhao[1]
*School of Aviation, Inner Mongolia University of Technology*
Hohhot, Inner Mongolia, China
huangyuanhao_work@163.com

Huang Shi[1]
*Institute of Earthquake Forecasting of the China Earthquake Administration*
Beijing, China
huangshi20@mails.ucas.ac.cn

Wang Hao
*School of Electrical Engineering, Yanshan University*
Qinhuangdao, Hebei Province, China
hwangwork@163.com

Meng Ruifeng*
*School of Aviation,
Inner Mongolia University of Technology*
Hohhot, Inner Mongolia, China
mrfnmgcn@imut.edu.cn

*Abstract*—This article introduces a multimodal motion planning (MMP) algorithm that combines three-dimensional (3-D) path planning and a DWA obstacle avoidance algorithm. The algorithms aim to plan the path and motion of obstacle-overcoming robots in complex unstructured scenes. A novel A-star algorithm is proposed to combine the characteristics of unstructured scenes and a strategy to switch it into a greedy best-first strategy algorithm. Meanwhile, the algorithm of path planning is integrated with the DWA algorithm so that the robot can perform local dynamic obstacle avoidance during the movement along the global planned path. Furthermore, when the proposed global path planning algorithm combines with the local obstacle avoidance algorithm, the robot can correct the path after obstacle avoidance and obstacle overcoming. The simulation experiments in a factory with several complex environments verified the feasibility and robustness of the algorithms. The algorithms can quickly generate a reasonable 3-D path for obstacle-overcoming robots and perform reliable local obstacle avoidance under the premise of considering the characteristics of the scene and motion obstacles.

## I. INTRODUCTION

Obstacle-overcoming robots have a wide range of applications in industry, planet exploration, inspection and security, medical service, agriculture, and so on [1–4]. In cities and factories, there are numerous unstructured obstacles such as steps and street edges. Because of their capability of obstacle-overcoming, the robots are more efficient as they move horizontally. Researchers have researched obstacle-overcoming robots and focused on how to improve the reliability and efficiency of robots in unstructured environments. On the one hand, the design and perception of the robot are the key points to detouring and overcoming obstacles. Previously, a novel obstacle-overcoming robot with a heterogeneous sensing system was presented [5]. The robot can overcome obstacles with deformable structures and a sensing system. However, limited by the impediments of a complex environment and motive obstacles such as cars and people, the robot mainly relies on the operator to send instructions for control. On the other hand, path planning and dynamic obstacle avoidance are critical technologies for robots to realize automated operations. Traditional 2-D and 3-D path planning algorithms [6–8] cannot be applied to the obstacle-overcoming robot. It is because the robot can only overcome parts of obstacles, the insurmountable limits the robot's motion. The primary method to achieve the goal and plan the path is bypassing all obstacles and generating an optimal path. The Chinese University of Hong Kong presented an Elastic band-based rapidly exploring random tree (EB-RRT) algorithm [9]. The algorithm, which combined the global planner and dynamic replanner, optimizes the heuristic trajectory efficiently to achieve real-time optimal motion planning and shows suitability and robustness in path planning tasks. Easwari Engineering College proposed a heuristic-based method combined with the simulated annealing algorithm-based approach for dynamic robot path planning [10].

The A-star algorithm is one of the most famous path planning algorithms and is mainly concerned with the operation time and the length of the generated path. Researchers continuously improved and derived numerous path planning methods for 2-D and 3-D space. The smooth A-star algorithm proposed by Universitas Gadjah Mada considers the above criteria. The algorithm smoothes the path to adapt to the tracking characteristics of the mobile robot [11]. 3-D A-star algorithms consume a lot of time and computing resources in space path planning tasks. The Federal University of Sao Carlos proposes a global planning algorithm that backtracks from target nodes and generates auxiliary nodes for unstructured indoor scenes [12]. In addition, for the refined application research of unstructured scenes, researchers used the terrain data map for planning experiments. Shanghai Ocean University optimized the A-star algorithm from two aspects of data structure and retrieval strategy and combined the high-precision data map to generate and verify the trajectory [13]. However, the method does not fully consider the vehicle's off-road performance and obstacle-surmounting ability.

The path planning algorithm provides a practical motion route reference for the robot. Nevertheless, there are unknown obstacles that are stationary or moving in authentic tasks. Therefore, the path planning algorithms are usually combined with the local obstacle avoidance algorithm. Such algorithms are particularly concerned with collision-free trajectory tracking [14]. Indian Institute of Technology Kanpur uses radial basis functions for parametric training of DMP motion planners. The algorithm achieves a good convergence effect

---

facing multi-objective static and dynamic obstacles [15]. Bionic strategies are a popular research direction for motion planning tasks. In tasks facing urban environments, pedestrians can be viewed as a dynamic obstacle with a large number and random motion. The University of British Columbia proposes a group surfing method to simulate the optimal walking group according to the laws of human motion. The algorithm is more secure and compatible than the traditional motion planning system [16]. Furthermore, the DWA algorithm is a dynamic obstacle avoidance strategy for high-speed moving robots. It fully considers the dynamics and search space of the robot, and has both higher computing speed and predicting precision [17. Xiamen University combined A-star and DWA algorithms into a hybrid path planning algorithm, which realizes simultaneous path tracking and obstacle avoidance [18].

As a multimodal intelligent vehicle, the obstacle-overcoming robot puts forward higher requirements for the autonomous decision-making ability of planning and obstacle avoidance algorithms. When facing complex unstructured scenes, how to perform motion accurately and efficiently is an urgent problem to be solved. This article reported a novel multimodal motion planning (MMP) algorithm, which combined 3-D path planning with obstacle avoidance. The robot can obtain higher searching efficiency and optimal path with the MMP algorithm. The main contributions are as follows.

(1) For path planning in unstructured environments, we presented a multimodal 3-D A-star algorithm with a novel heuristic function and introduced the obstacle-overcoming cost value into the heuristic function, which is the basis for total cost calculation and decision-making.

(2) A decision strategy is reported to switch between A-star and greedy best-first strategy (GBFS) search approaches based on whether the obstacle can be overcome.

(3) The multimodal A-star algorithm is integrated with the DWA, and the multimodal motion planning (MMP) strategy for the obstacle-overcoming robots is obtained.

The rest of this paper is organized as follows. Section II demonstrates the strategy and principle of the multimodal 3-D A-star algorithm. Section III reported the strategy of MMP, which combined the path planning algorithm with DWA, and exhibited the performance of the MMP algorithm in specific scenes. In section IV, we set up synthetic scenarios to validate the algorithm and discuss the results. Finally, conclusions are summarized in Section V.

## II. MULTIMODAL 3-D A-STAR ALGORITHM

In this section, we demonstrate a novel multimodal 3-D A-star Algorithm considering the cost of obstacle-overcoming. Firstly, Section II.A introduces the heuristic search principle of the traditional A-star algorithm, focusing on the design of its heuristic function. Subsequently, Section II.B demonstrates how we introduce the obstacle of overcoming cost and unify it with the distance cost, forming a new heuristic function. It plans scenarios including flat roads, surmountable obstacles, and insurmountable obstacles. The features reflect the ability to switch between modes of movement and obstacle overcoming and to plan three-dimensional scenes. In Section II.C, we demonstrated and commented on the algorithm in specific obstacle-overcoming scenarios.

| Algorithm 1 Traditional A-star algorithm |
|---|
| 1: Make P[*start*] as *open list*. |
| 2: **while** *open list* ≠ *empty* **do** |
| 3:   select the node **P[i]** from the *open list* whose value of evaluation function **F(P[i])** is smallest. |
| 4:   Make *P [i]* as *close list* |
| 5:   **if** *P [i] =p[end]* **then** |
| 6:     return "path is found" |
| 7:   **else** |
| 8:     Select the successor node $P_i[j]$ around the node $P[i]$ |
| 9:     //calculate $F(P_i[j])$ |
| 10:    H ← Calculate the heuristic search cost by the Manhattan distance from the target point. |
| 11:    G ← The cost of moving to the current position + the next move cost. |
| 12:    F = H + G |
| 13:    **if** $P_i$ belongs to *obstacle* or *close list* node **then** |
| 14:      continue; |
| 15:    **end if** |
| 16:    Mark $P_i[j]$ as *open list*. |
| 17:    **if** $P[j]$ belongs to *open list* and $F(P_i[j]) < F(P_m[j])$ when $P[m]$ was marked as *close list* **then** |
| 18:      Set parent node of $P[j]$, $F(P[i] = F(P_i[j])$. |
| 19:    **end if** |
| 20:  **end if** |
| 21: **end while** |
| 22: **return** "the path cannot be found" |

### A. Traditional A-star Algorithm

The A-star algorithm is a classic path planning algorithm proposed by Hart et al. in 1968 [19]. The algorithm obtains the minimum cost path by calculating the minimum cost, such as time, distance, risk, and other factors. Algorithm 1 represents the pseudocode of the traditional A-star algorithm, and its core lies in calculating the total cost F. Total cost F is calculated by heuristic search cost $H$ and the sum cost to reach the current position, and the next grid $G$. $H$ is based on the Manhattan distance from the target point. At this time, $P[i]$ and $P_i[j]$ represent the current node and the nodes around the node $P[i]$, respectively. Nodes in *openlist* represent reachable and unreachable nodes. The total cost $F$ of grid $i$ is evaluated as

$$F(P[i]) = H(P[i]) + G(P[i]) \qquad (1)$$

where $H(P[i])$ is the cost of the path to the aim position $P[i]$. $G(P[i])$ is the cost of the path from a position of start to position grid $i$. When A-star performs path planning, the map will be rasterized and used to calculate. The algorithm expands reasonable nodes through search strategies. Common search approaches include A-star best-first strategy (ABFS) search and GBFS search. The traditional A-star algorithm uses a special heuristic search function to expand and use Euclidean distance as a heuristic function to prevent falling into an optimal solution.

### B. Heuristic Function Considering Obstacle-overcoming

As a 2-D search algorithm, A-star has fast operation speed and optimal solution characteristics when using a good heuristic function. In order to expand the least number of nodes when searching for the optimal path, the search algorithm needs to decide which node to expand to continuously. The A-star algorithm uses a heuristic and efficient function to evaluate optional nodes. When the algorithm encounters an obstacle, the grid at the obstacle is not considered for the next node. The search continues along the obstacle and outside.

In order to consider the possibility of an obstacle as the next node during the search path process, we introduce a new variable of the obstacle-overcoming cost $T$ as a heuristic function. The main cost incurred by the robot in surmounting obstacles is the time cost. When quantifying the obstacle clearance cost, we mainly consider the robot's speed and obstacle clearance time. $T$ is a time parameter that considers the performance of the obstacle-surmounting robot, which is divided into the climbing obstacle time $T_u$ and the descending obstacle time $T_d$. Let $T(i)$ and $T(j)$ denote the cost of the current grid and its surrounding grids, respectively. At this point, we can write $T(i)$ as the sum of two parts.

$$T(i) = t_c(i) + h(i) \cdot t_o \quad (2)$$

where $t_o$ is the cost for the robot to lift or climb over the unit height at the grid of $i$. $h(i)$ is the height of the obstacle at the grid of $i$. The threshold also needs to be set according to the robot's ability to overcome obstacles. When the obstacle height exceeds the maximum obstacle clearance height of the robot, the obstacle is regarded as an insurmountable obstacle and does not participate in the expansion calculation of the next grid. When the obstacle height is less than or equal to the maximum obstacle clearance height of the robot, it needs to participate in the calculation.

The heuristic function is a crucial part of heuristic search. Traditional algorithms use Euclidean distance or Manhattan distance as a heuristic function. However, the parameter $T$ introduced in this article is based on time, and the robot's speed will significantly impact the decision to overcome obstacles or detours. Therefore, we use the robot's speed as a parameter to participate in the cost calculation and convert the Manhattan distance function into a time cost:

$$H_t(i) = \frac{\sum_{i}^{n} d(x_i, y_i)}{v} \quad (3)$$

$$G_t(i, j) = G(i) + \frac{c(j)}{v} \quad (4)$$

where $H_t(i)$ is the time cost of the path to the aim position, and $G_t(i, j)$ is the actual time cost of an optimal path from the outset to $P(i)$. We obtain the evaluation function $F_t(i)$ as follows.

$$F_t(i) = \frac{\sum_{i}^{n} d(x_i, y_i)}{v} + [G(i) + \frac{c(j)}{v}] + [t_c(i) + h(i) \cdot t_o] \quad (5)$$

**Algorithm 2** New heuristic function for obstacle crossing cost
1: $H_i[j] \leftarrow$ Calculate the height difference between $P[i]$ and $P_i[j]$
2: $Tc_i[j] \leftarrow$ Calculate the heuristic search cost by the Manhattan distance from $P[i]$ to $P_i[j]$
3: **If** $H_i[j] >$ Maximum Obstacle Clearance Height **then**
4:     **return** "Unreachable"
5: **else if** $H_i[j] <$ Maximum direct passing Height **then**
6:     $T_i[j] \leftarrow Tc_i[j]$
7: **else**
8:     $T_i[j] \leftarrow T_0 \times H_i[j] + Tc_i[j]$
    **end if**

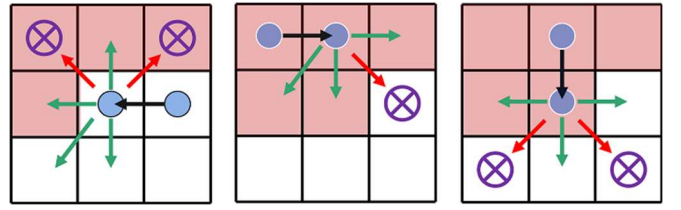

Fig. 1. The choice of quadtree and octree index methods for special obstacle-overcoming nodes.

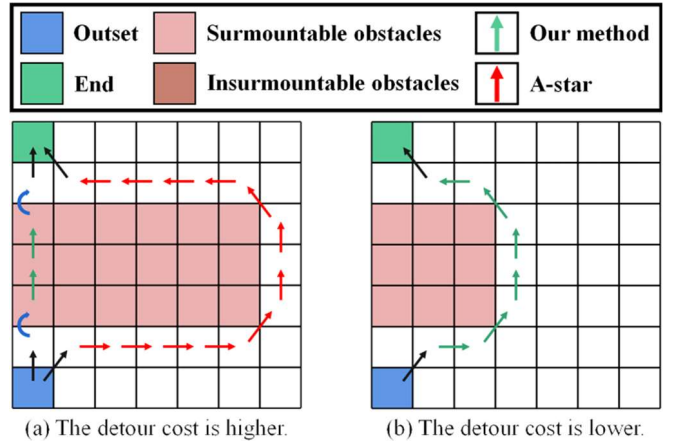

(a) The detour cost is higher.  (b) The detour cost is lower.

Fig. 2. Comparing the multimodal 3-D A-star algorithm with the traditional A-star algorithm. (a) The cost of detouring is higher than overcoming an obstacle. (b) The cost of detouring is lower than overcoming an obstacle.

It is easy to establish a time cost, considering the velocity and ability of obstacle-overcoming. Let $F'$ estimate the value of $F_t(i)$, and the algorithm can choose the smallest cost for $F'$. This implies $F' \geq F_t(i)$.

Strategy 2 represents the pseudocode after introducing the obstacle-overcoming cost. During the movement process, the obstacle-overcoming robot moves in any direction. When overcoming an obstacle, it is usually necessary to move directly against the obstacle. Therefore, in the planning task of unstructured scenes, in addition to introducing new parameters and uniting the heuristic function, the indexes methods of different grids are also optimized. Figure 1 shows the search for three critical nodes of obstacle overcoming. Regardless of whether it is an upper obstacle or a lower obstacle, the algorithm uses the hybrid indexing method of quadtree and octree when searching for the next node. The node overcoming the obstacle uses the quadtree index, and the node that does not overcome the obstacle uses the octree index.

A multimodal 3-D A-star algorithm is obtained, which considers the cost of obstacle overcoming and unifies the quantification standard of the total cost. Figures 1 to 3 illustrate the difference between the proposed algorithm and the traditional A-star search algorithm in obstacle scenarios. In the schematic diagram, we set up 6 maps with their characteristics. The maps contain the outset, the ending point, the surmountable obstacles, and insurmountable obstacles.

To illustrate the algorithm's feasibility, we set the cost of moving a single grid as 1, the cost of climbing up the obstacle is 4, and the cost of climbing down the obstacle is 3. The specific cost estimation needs to refer to the estimation method in Section II.A and the actual physical parameters of the obstacle-overcoming robot.

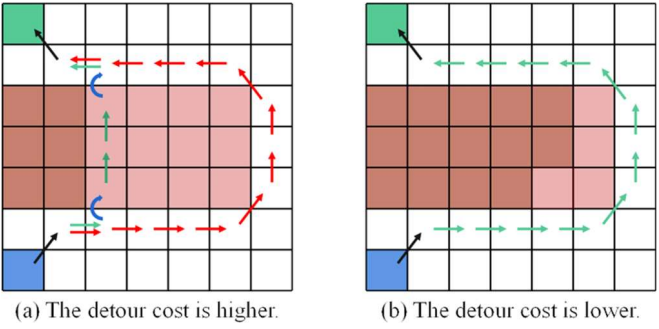

Fig. 3. Scenes with flat roads, surmountable and insurmountable obstacles. (a) The cost of detouring is higher than overcoming obstacle. (b) The cost of detouring is lower than overcoming obstacle.

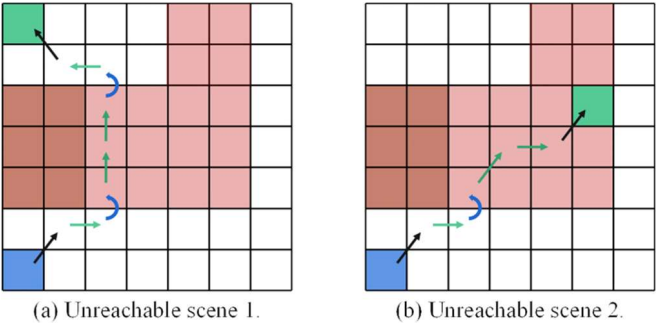

Fig. 4. Endpoints that cannot be reached by 2-D path planning algorithms.

Figure 2 shows two different situations in which there are surmountable obstacles. And when the cost of detouring the obstacle is higher than overcoming the obstacle. The proposed algorithm selects a reasonable place to overcome the obstacle and reach the end as shown in Fig. 2 (a). The traditional A-star algorithm does not have the conditions to consider the obstacle. When the detour cost is lower than the obstacle-overcoming cost, the algorithm obtains the optimal path that is consistent with the traditional algorithm as shown in Fig. 2 (b).

When the scene contains both flat roads, surmountable obstacles, and non-traversable obstacles, the algorithm does not consider the non-traversable obstacles as the next expansion area. Similar to Fig. 2, Fig. 3 shows path planning results with different detour costs. Figures 3(a) and (b) show that the detouring cost is higher or lower than the obstacle overcoming cost. The difference from Fig. 1 is that the robot calculated the cost accurately and reasonably after detouring the insurmountable obstacle.

In addition, as a multimodal 3-D path planning algorithm, it can also navigate to the endpoints that the 2-D path planning algorithm cannot reach. Figure 4(a) depicts a path that bypasses the insurmountable obstacle and then climbs over the surmountable obstacle to reach the endpoint. Figure 4(b) depicts a path reaching the endpoint of a surmountable obstacle.

### C. Decision-making for Switching Search Approaches

It has been proved that for a suitable heuristic function, the A-star algorithm can obtain the globally optimal path from the starting point to any non-closed node n. In a 2-D scene, ABFS is an approach that considers both the speed of operation and the consequence. However, in unstructured scenes, the path has several obvious features. For example, there may be passages or slopes next to high-cost surmountable obstacles (the cost is the same as the flat road). On the one hand, ABFS search in unstructured scenarios consumes more computing time and resources than the GBFS search. Because the algorithm search for more redundant grids to make sure that the path is the best one. On the other hand, GBFS are prone to fall into the trap of locally optimal solutions in specific scenarios. Considering the characteristics of unstructured scenes and the acquisition of the optimal global solution, we propose a decision-making strategy for switching search methods.

Figure 5 demonstrates three different searching approaches' performance in the same grid map. The greedy method is the shortest path search method to select the optimal local solution for each grid expansion. Unlike Eq. (5), the evaluation function used by the greedy method ignores the actual time cost of an optimal path from the outset to the current grid. That means that although the number of search grids of the GBFS is smaller than the ABFS, it is also possible to fall into a locally optimal solution. Meanwhile, all searched grid information is stored in an open list for reading and calculation. Figure 5 (a) shows the GBFS for searching grids and generating paths in a map. It is not an optimal path caused by the distance-priority characteristic of the evaluation function. The generated path and the positions of grids searched by the ABFS are shown in Fig. 5 (b). Although the method searches more grids than the GBFS, the optimal path for this map is obtained. In order to ensure that the search speed can be improved as much as possible based on obtaining the optimal global solution, we combine the above two search methods. GBFS. The greedy method searches along the barrier until it GBFS. The greedy method searches along the barrier until it finds a lower or equal cost to the switching grids and stops. That is, when finding a jump point with a shorter heuristic distance to the vertical direction of the obstacle extension, it switchs back to the ABFS. In addition, a jump search method is introduced to optimize the jump point obtained by the greedy method.

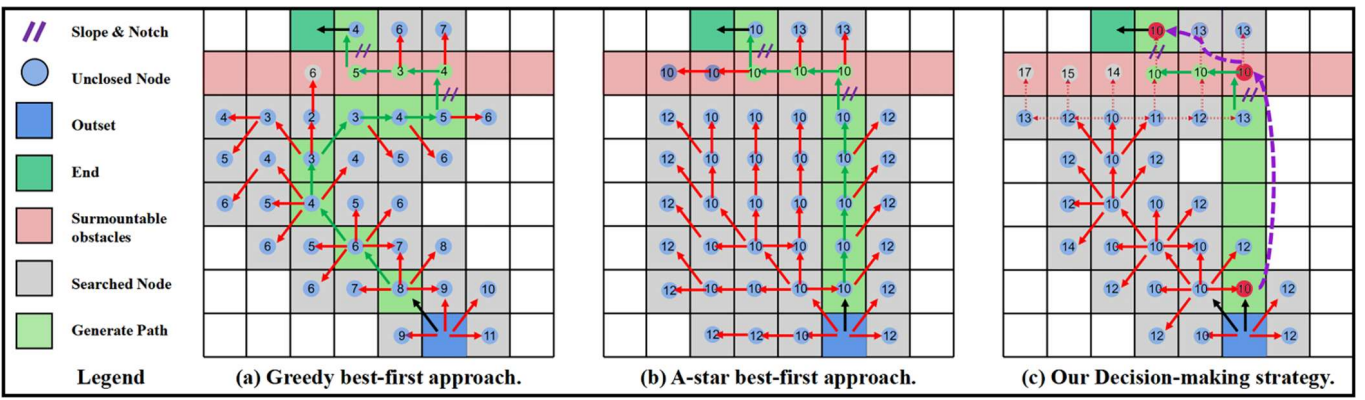

Fig. 5. In the unstructured scene, our searching algorithm is compared with the ABFS and GBFS.

As shown in Fig. 5 (c), when the algorithm searched a grid with a barrier edge cost of 10, the search method is switched to the greedy method. The search path of the GBFS is represented by the red dashed line, and a path-first search is performed along the edge of the obstacle. When the algorithm finds a grid with a cost of 10, we find critical grids with lower cost and the greedy search ends. The grid is a jump point for the jump search. The next jump point can be obtained in the same way. At this time, the results of the jump search can be obtained by the purple dotted line. Only the three key nodes are counted in the search grid number. The jump point algorithm meets the following rule. Only the jump point is put into the close list. The final path points are also jump points [20].

Table I shows the critical parameters comparison of the algorithms in the example. It is assumed that each movement of a grid is a unit distance. Thus, the proposed strategy combined with the hop algorithm uses the least number of search grids to obtain the optimal path. However, if the problem mentioned in the above rules occurs, the strategy takes more time to search but is still lower than the single ABFS.

TABLE I. CRITICAL PARAMETERS COMPARISON

|  | Greedy | A-star | Our strategy |
|---|---|---|---|
| Grids | 37 | 42 | 35 |
| Distance | 16 | 10 | 10 |

## III. MULTIMODAL MOTION PLANNING ALGORITHM

The dynamic window method is a local dynamic obstacle avoidance algorithm. It performs sampling in a limited space according to the acceleration and deceleration performance of the robot. The movement trajectory of the robot based on the speeds is simulated and evaluated to select a good trajectory for movement [21]. The current mainstream motion control methods are suitable for spatiotemporal methods. They pass through the trajectory of known obstacles, assuming that a moving obstacle will be stationary in the future. The University of Zagreb reported the integration of a dynamic window and path planning module [22,23]. The algorithm incorporated both target heading $\vartheta_{head}(v,\omega)$ and linear velocity $\vartheta_{vel}(v,\omega)$ ensures a single path alignment measure $\vartheta_{path}(v,\omega)$. The global objective function can be expressed as

$$\Gamma(v,\omega) = \lambda \vartheta_{clear} + (1-\lambda)\vartheta_{path} \qquad (6)$$

The path-tracing criterion requires that the possible motion trajectories of the robot are associated with local machines and path configurations. This improves the robot's translation speed constraints and clarifies the heading in the local window. The tracking method described above is used for orientation confirmation. However, the obstacles in the scene need to be overcome, detected, and avoided by the DWA algorithm. Thus, we propose a simple two-step scene overlay method for motion planning. Figure 6 shows the concept of the MMP algorithm. The robot obtains the planning path in the first step as shown in Fig. 6(a). Then, the robot moves in the environment with motive and still obstacles. Combining the planning path and the strategy of DWA by overlaying the two maps. Finally, the robot detours the obstacles in step two but ignores the surmountable obstacles in the first step as shown in Fig. 6(d).

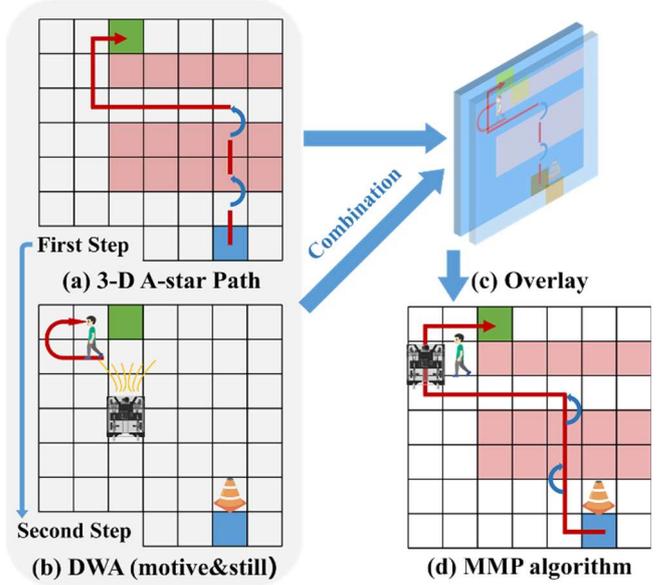

Fig. 6. Multimodal motion planning (MMP) algorithm combining 3-D A-star path planning and DWA algorithm.

## IV. EXPERIMENTS AND DISCUSSION

We build a map abstracted from a factory campus to validate the multimodal motion planning strategy proposed in this study. The scene includes flat roads, gentle slopes regarded as flat roads, road steps that can be crossed, and scenes that cannot be crossed. Through the motion simulation of the obstacle-overcoming robot in the above scenario, the motion planning performance of the algorithm in the comprehensive scenario is tested. We focus on the following two key points.

(1) Whether the path generated by the multimodal 3-D A-star path planning algorithm is reasonable, and whether the operation time is optimized compared with the traditional strategy.

(2) Does the DWA track the planned trajectory well and respond reasonably to unknown obstacles?

An experimental run is presented in Fig. 7. The simplified scene settings for the factory map are shown in Fig.7 (a). We used different search strategies for path planning through the proposed algorithm. The path of generation of the strategy and GBFS are shown in Figs. 7(b-1) and 7 (b-2), respectively. The green area is the range of grid search, and the red line represents the final generated path. The search strategy of this article and ABFS obtain the exact optimal path to the endpoint, while the GBFS chooses a different path. The optimal path reaches the end point after two obstacle crossings and one ramp crossing. The GBFS's path falls into the scene obstacle's local trap in Fig.7(a). In terms of search time, GBFS only takes 0.07 s, while ABFS takes 0.61 s. The algorithm in this paper takes 0.54s, which is only 11.5% higher than ABFS.

Figure 7(c) shows the MMP algorithm's performance in the scene. The red line in Fig. 7(c-1) demonstrates the path, which is generated by the algorithm. The robot rams the unknown obstacle without the DWA algorithm shown in Fig.7 (c-2).

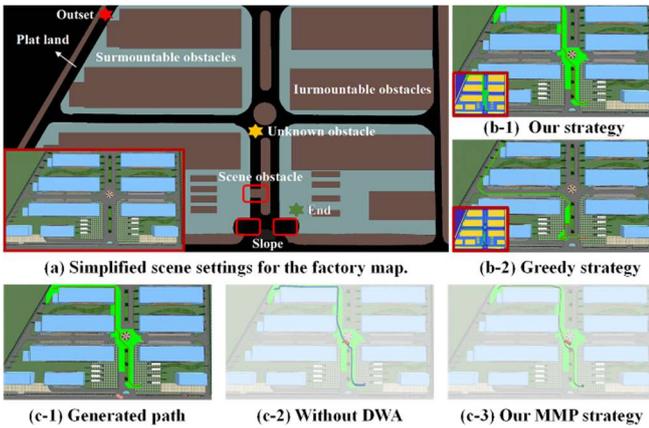

Fig. 7. Comprehensive 3-D path planning experiments and comparisons in complex factory environments.

Through the MMP strategy, the robot detoured the unknown obstacle to achieve the endpoint. In addition, we also set random starting points and obstacles for experiments. During the process, we found that DWA sometimes fails to achieve the desired effect with fixed parameters, and the robustness is a little bit poor.

## V. CONCLUSION AND FUTURE WORK

A novel multimodal 3-D A-star path planning algorithm is proposed to introduce the obstacle-overcoming cost and unified evaluation function. The algorithm plans the path in environments with flat land, surmountable obstacles, and insurmountable obstacles. In addition, we expressed a decision-making strategy between ABFS and GBFS. The strategy considers the features of unstructured environments and obtains the optimal path as fast as possible. In the illustrated examples and experiments, the optimal path can always be obtained in less time than A-star. Owing to the open list reducing the time spent reading grid data, the time cost of the proposed strategy is lower than A-star convenience in search time but slightly higher than the greedy method. Finally, we introduced the multimodal motion plan strategy and expert with the map that abstracted outdoor scenes and verified the feasibility and advantages of the algorithm. Although the algorithm has achieved good results in the simulation experiments, the fusion of the decision strategy and the jump point algorithm only considers the excellent performance in simple scenarios. In future work, we will combine this algorithm to develop a pruning rule to optimize the grid expansion strategy under obstacle overcoming conditions. At the same time, designing a local planning algorithm to solve the low robustness problem of the DWA algorithm is also under our consideration.

ACKNOWLEDGMENT

The authors wish to express their gratitude to the Inner Mongolia Autonomous Region Intellectual Property Special Project for their funds.